\documentclass[runningheads]{llncs}
\usepackage{hyperref}

\usepackage{amsmath}
\usepackage{amsfonts}
\usepackage{amssymb}
\usepackage{float}
\usepackage{lipsum}
\usepackage{wrapfig}
\usepackage{multicol}
\usepackage[table, dvipsnames]{xcolor}
\usepackage[square,numbers,sort&compress]{natbib}
\usepackage{tabulary}
\usepackage{enumitem}
\usepackage{graphicx}
\usepackage{subfig}
\usepackage{caption}
\usepackage[strings]{underscore}
\usepackage{algorithm}
\usepackage[noend]{algpseudocode}
\usepackage{bbm}
\usepackage{gensymb}
\usepackage{mathtools}
\usepackage{xspace}

\newlist{inlist}{enumerate*}{1}
\setlist[inlist]{label=(\arabic*)}
\newlist{inlistalpha}{enumerate*}{1}
\setlist[inlistalpha]{label=(\alph*)}

\newcolumntype{L}[1]{>{\raggedright\arraybackslash}p{#1}}
\newcolumntype{C}[1]{>{\centering\arraybackslash}p{#1}}

\newcommand{\R}{\ensuremath{\mathbb{R}}}

\newcommand{\para}[1]{\noindent\textbf{#1}.~~}

\newcommand{\raypoint}{\mathbf{p}}
\newcommand{\raydir}{\hat{\mathbf{r}}}

\newcommand{\wiretip}{\mathbf{x}}
\newcommand{\wiredir}{\hat{\mathbf{v}}}
\newcommand{\skillfactor}{\lambda_{\rm adj}}
\newcommand{\clip}[3]{\mathrm{clip}\hspace{-1.5pt}\left(#1,\,#2,\,#3\right)}

\newcommand{\sourcedetector}{d_{\rm sd}}
\newcommand{\sourcepoint}{d_{\rm sp}}
\newcommand{\sensor}{w_{\rm s}}

\newcommand{\rot}[2]{\mathrm{Rot}\left(#1,\,#2\right)} 
\newcommand{\poswire}{\texttt{position-wire}\xspace}
\newcommand{\insertwire}{\texttt{insert-wire}\xspace}
\newcommand{\insertscrew}{\texttt{insert-screw}\xspace}
\newcommand{\hunting}{\texttt{hunting}\xspace}
\newcommand{\assessment}{\texttt{assessment}\xspace}
\newcommand{\ap}{\texttt{AP}\xspace}
\newcommand{\lateral}{\texttt{lateral}\xspace}

\DeclareMathOperator{\uniform}{\mathcal{U}}
\DeclareMathOperator{\unifangle}{\mathcal{U}_{\measuredangle}}
\DeclareMathOperator{\unifsphere}{\mathcal{U}_{\circ}}

\author{Benjamin~D.~Killeen\inst{1} \and
Han Zhang\inst{1}\and
Jan Mangulabnan\inst{1}\and
Mehran Armand\inst{2}\and
Russell H.~Taylor\inst{1}\and
Greg Osgood\inst{2}\and
Mathias Unberath\inst{1}
}
\authorrunning{B.D.~Killeen et al.}

\institute{Johns Hopkins University, Baltimore, MD, USA\\
\inst{1}\email{\{killeen, hzhan206, jmangul1, rht, unberath\}@jhu.edu}\\
\inst{2}\email{\{marmand2, gosgood2\}@jhmi.edu}
}

\begin{document}

\title{Pelphix: Surgical Phase Recognition from X-ray Images in Percutaneous Pelvic Fixation}
\titlerunning{Pelphix: Surgical Phase Recognition from X-ray Images}



\maketitle              

\begin{abstract}

Surgical phase recognition (SPR) is a crucial element in the digital transformation of the modern operating theater. While SPR based on video sources is well-established, incorporation of interventional X-ray sequences has not yet been explored.
This paper presents Pelphix, a first approach to SPR for X-ray-guided percutaneous pelvic fracture fixation, which models the procedure at four levels of granularity -- corridor, activity, view, and frame value -- simulating the pelvic fracture fixation workflow as a Markov process to provide fully annotated training data. 
Using added supervision from detection of bony corridors, tools, and anatomy, we learn image representations that are fed into a transformer model to regress surgical phases at the four granularity levels. 
Our approach demonstrates the feasibility of X-ray-based SPR, achieving an average accuracy of 93.8\% on simulated sequences and 67.57\% in cadaver across all granularity levels, with up to 88\% accuracy for the target corridor in real data. This work constitutes the first step toward SPR for the X-ray domain, establishing an approach to categorizing phases in X-ray-guided surgery, simulating realistic image sequences to enable machine learning model development, and demonstrating that this approach is feasible for the analysis of real procedures. As X-ray-based SPR continues to mature, it will benefit procedures in orthopedic surgery, angiography, and interventional radiology by equipping intelligent surgical systems with situational awareness in the operating room.\footnote{
Code and data available at \url{https://github.com/benjamindkilleen/pelphix}.}

\end{abstract}
\section{Introduction}

In some ways, surgical data is like the expanding universe: 95\% of it is dark and unobservable \cite{Caldwell2009Apr}. The vast majority of intra-operative X-ray images, for example, are ``dark'', in that they are not further analyzed to gain quantitative insights into routine practice, simply because the human-hours required would drastically outweigh the benefits. As a consequence, much of this data not only goes un-analyzed but is discarded directly from the imaging modality after inspection. Fortunately, machine learning algorithms for automated intra-operative image analysis are emerging as an opportunity to leverage these data streams. A popular application is surgical phase recognition (SPR), a way to obtain quantitative analysis of surgical workflows and equip automated systems with situational awareness in the operating room (OR). SPR can inform estimates of surgery duration to maximize OR throughput \cite{Guedon2021Nov} and augment intelligent surgical systems, \emph{e.g.}~for suturing \cite{Varier2020Aug} or image acquisition \cite{DaCol2021Jan, Kausch2021Sep, Killeen2023Jun}, enabling smooth transitions from one specialized subsystem to the next. Finally, SPR provides the backbone for automated skill analysis to produce immediate, granular feedback based on a specific surgeon's performance \cite{Wu2021May, DiPietro2019Nov}.

\begin{figure}[t]
  \centering
  \includegraphics[width=\columnwidth]{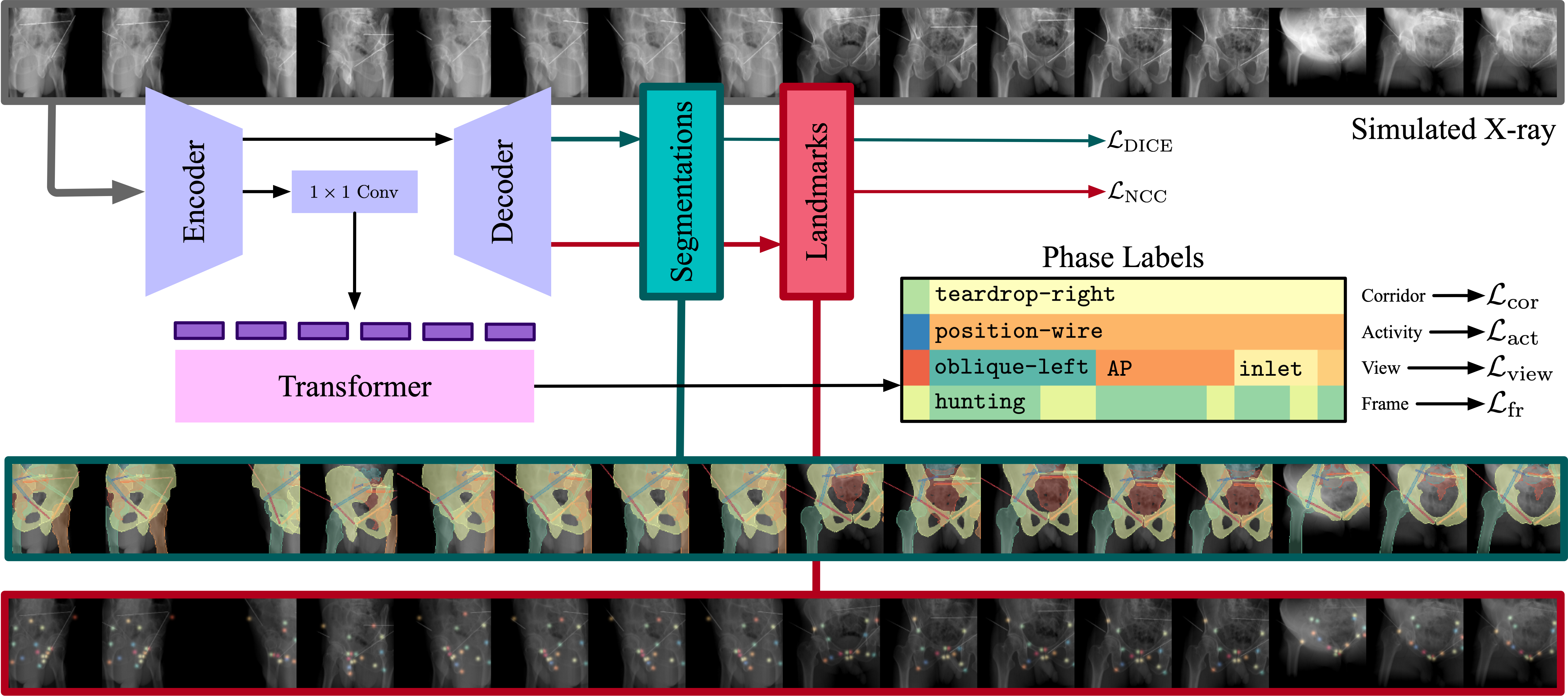}
  \caption{
    Our model architecture incorporates frame-level spatial annotations using a U-Net encoder-decoder variant. Anatomical landmarks and segmentation maps provide added supervision to the image encoder for a transformer, which predicts the surgical phase. The images shown here are the result of Markov-based simulation of percutaneous fixation, used for training.}
  \label{fig:architecture}
\end{figure}

The possibilities described above have motivated the development of algorithms for surgical phase recognition based on the various video sources in the OR \cite{Padoy2019Mar, Zhang2022Jul, Valderrama2022, Zisimopoulos2018Sep}. However, surgical phase recognition based on interventional X-ray sequences remains largely unexplored. Although X-ray guidance informs more than 17 million procedures across the United States (as of 2006) \cite{Kim2012Jul}, the unique challenges of processing X-ray sequences compared to visible or structured light imaging have so far hindered research in this area. Video cameras collect many images per second from relatively stationary viewpoints. By contrast, C-arm X-ray imaging often features consecutive images from vastly different viewpoints, resulting in highly varied object appearance due to the transmissive nature of X-rays. 
X-ray images are also acquired irregularly, usually amounting to several hundred frames in a procedure of several hours, limiting the availability of training data for machine learning algorithms.

Following recent work that enables sim-to-real transfer in the X-ray domain \cite{GaoSyntheX}, we now have the capability to train generalizable deep neural networks (DNNs) using simulated images, where rich annotations are freely available. \emph{This paper represents the first step in breaking open SPR for the X-ray domain, establishing an approach to categorizing phases, simulating realistic image sequences, and analyzing real procedures.} We focus our efforts on percutaneous pelvic fracture fixation, which involves the acquisition of standard views and the alignment of Kirschner wires (K-wires) and orthopedic screws with bony corridors \cite{Simonian1994}. We model the procedure at four levels, the current target corridor, activity (\poswire, \insertwire, and \insertscrew), C-arm view (\ap, \lateral, etc.), and frame-level clinical value.
Because of radiation exposure for both patients and clinicians, it is relevant to determine which X-ray images are acquired in the process of ``fluoro-hunting'' (\hunting) versus those used for clinical \assessment. Each of these levels is modeled as a Markov process in a stochastic simulation, which provides fully annotated training data for a transformer architecture.

\section{Related Work}
\label{sec:related-work}

SPR from video sources is a popular topic, and has benefited from the advent of transformer architectures for analyzing image sequences. The use of convolutional layers as an image encoder has proven effective for recognizing surgical phases in endoscopic video \cite{Zhang2022Jul}, laparoscopic video \cite{Valderrama2022}, and external time-of-flight cameras \cite{Czempiel2021Sep}. These works especially demonstrate the effectiveness of transformers for dealing with long image sequences \cite{Czempiel2021Sep}, while added spatial annotations improve both the precision and information provided by phase recognition \cite{Valderrama2022}. Although some work explores activity recognition in orthopedic procedures \cite{Kadkhodamohammadi2021May,Hossain2018Jun} they rely on head-mounted cameras with no way to assess tool-to-tissue relationships in percutaneous procedures. The inclusion of X-ray image data in this space recenters phase recognition on patient-centric data and makes possible the recognition of surgical phases which are otherwise invisible.


\section{Method}

The Pelphix pipeline consists of stochastic simulation of X-ray image sequences, based on a large database of annotated CT images, and a transformer architecture for phase recognition with additional task-aware supervision. A statistical shape model is used to propagate landmark and corridor annotations over 337 CTs (see supplement), as shown in Fig.~\ref{fig:annotations}. The simulation proceeds by randomly aligning virtual K-wires and screws with the annotated corridors (Section~\ref{sec:image-seq}). In Section~\ref{sec:architecture}, we describe a transformer architecture with a U-Net style encoder-decoder structure enables sim-to-real transfer for SPR in X-ray.

\label{sec:ct-img-ann}

\begin{figure}[t]
    \centering
    \subfloat[\scriptsize Corridors and Landmarks]{
    \includegraphics[width=0.4\columnwidth]{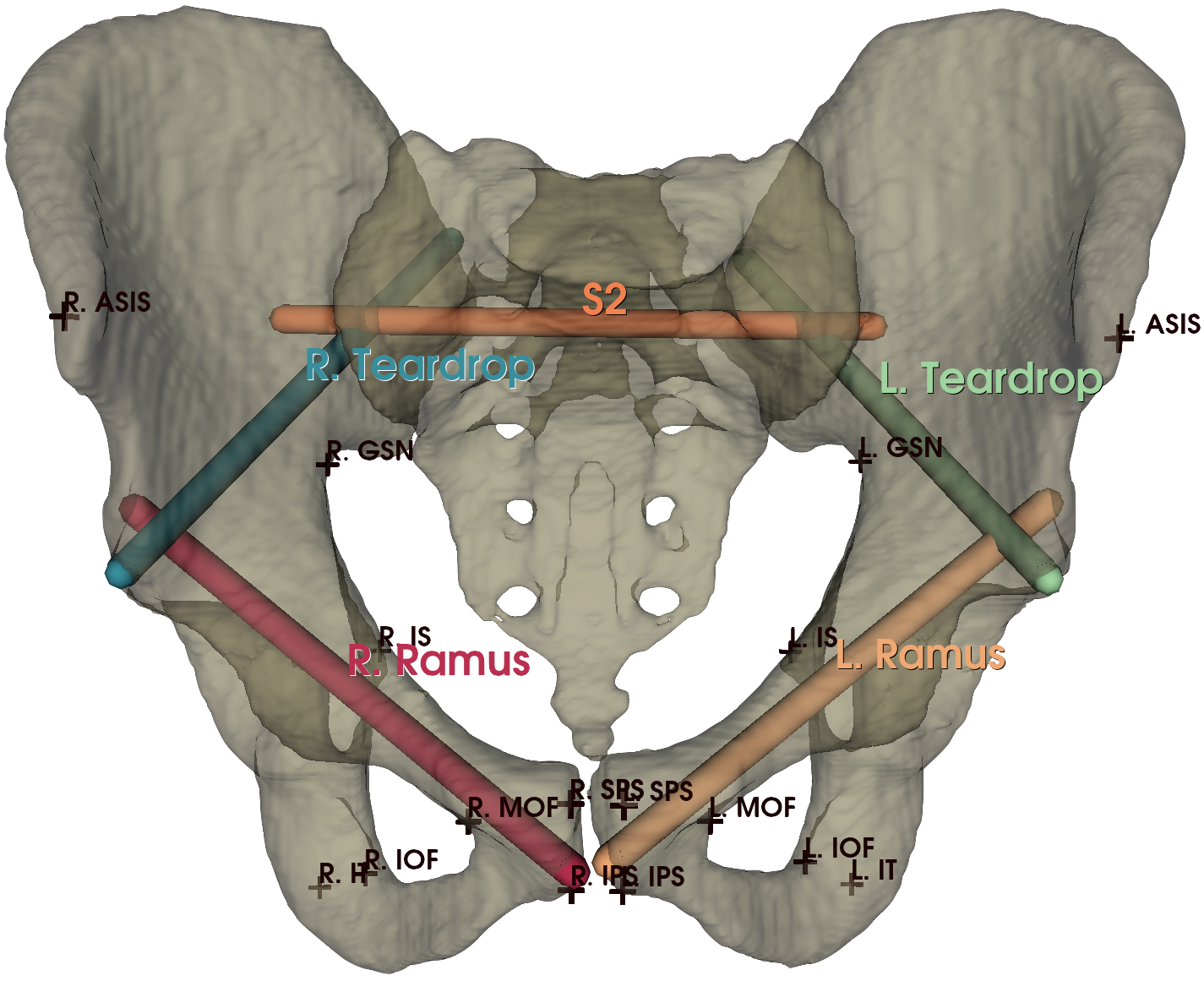}
    \label{fig:annotations}
    }
    \subfloat[\scriptsize Anterior Pelvic Plane]{
    \includegraphics[width=0.4\columnwidth]{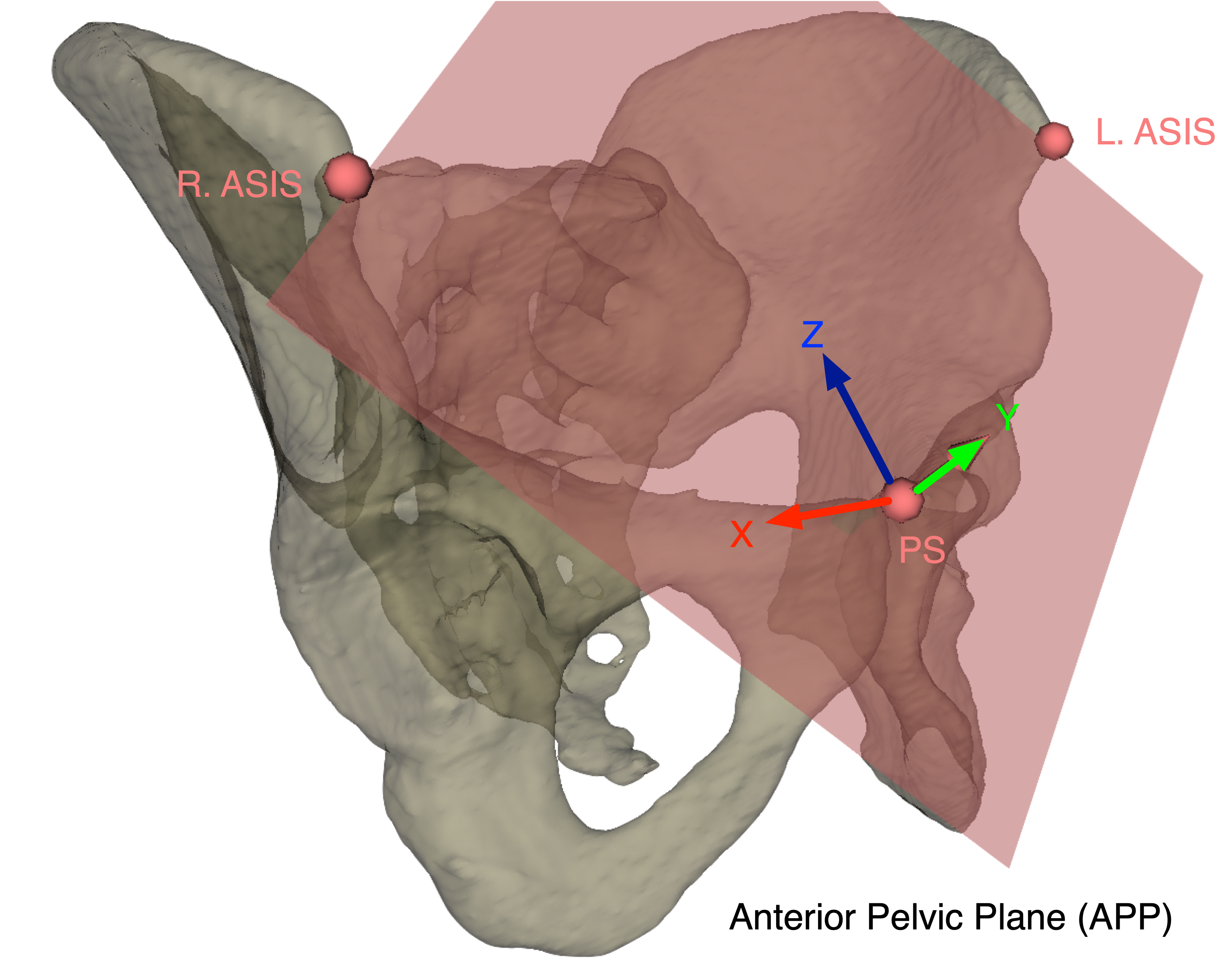}
    \label{fig:app-plane}
    }
    \caption{\protect\subref{fig:annotations} The ramus, teardrop and S2 bony corridors, as well as 16 anatomical landmarks with added supervision for phase recognition. 
    \protect\subref{fig:app-plane} The anterior pelvic plane (APP) coordinate system is used to define principle ray directions for standard views of the pelvis, enabling realistic simulation of image sequences for percutaneous fixation.}
    \label{fig:overview}
\end{figure}

\subsection{Image Sequence Simulation for Percutaneous Fixation}
\label{sec:image-seq}

Unlike sequences collected from real surgery \cite{Padoy2019Mar} or human-driven simulation \cite{munawar2021virtual}, our workflow simulator must capture the procedural workflow while also maintaining enough variation to allow algorithms to generalize. We accomplish this by modeling the procedural state as a Markov process, in which the transitions depend on evaluations of the projected state, as well as an adjustment factor $\skillfactor \in [0,1]$ that affects the number of images required for a given task. A low adjustment factor decreases the probability of excess acquisitions for the simulated procedure. In our experiments, we sample $\skillfactor \in \uniform(0.6, 0.8)$ at the beginning of each sequence.

\begin{figure}[t]
    \centering
    \includegraphics[width=\columnwidth]{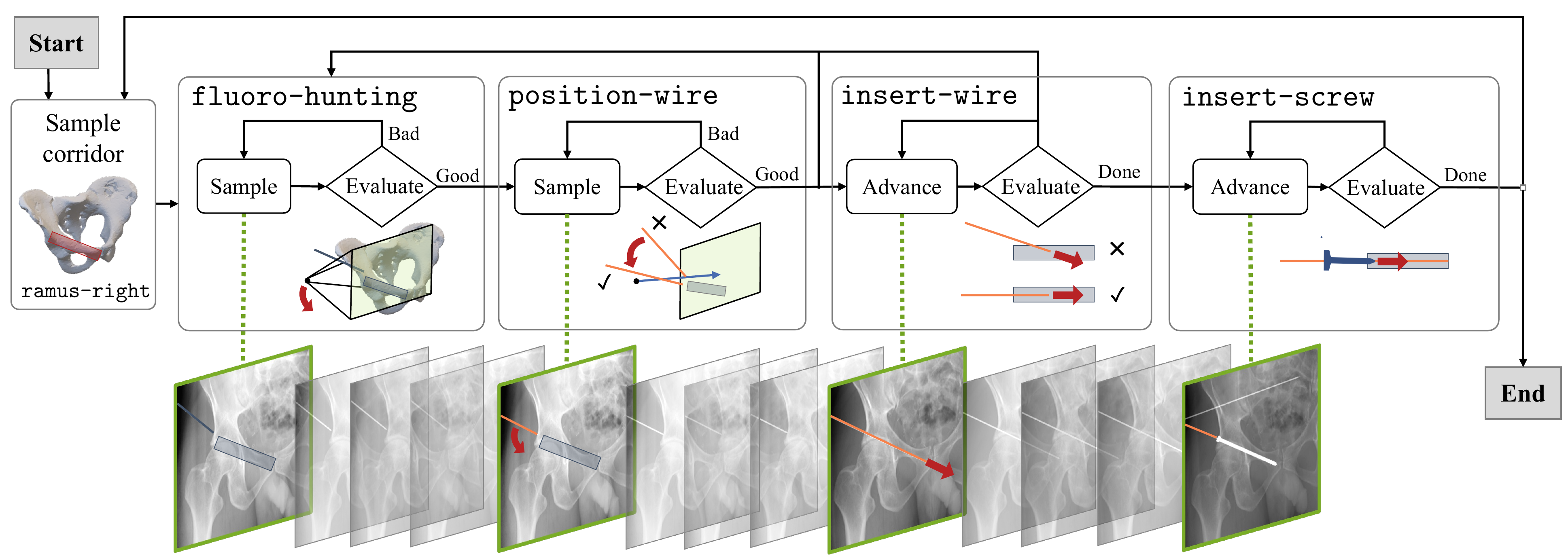}
    \caption{The image sequence simulation pipeline for Pelphix. We model the procedure as a Markov random process, where transition probabilities depend on realistic evaluation of the current frame.}
    \label{fig:simulation-overview}
\end{figure}

Fig.~\ref{fig:simulation-overview} provides an overview of this process. Given a CT image with annotated corridors, we first sample a target corridor with start and endpoints $\mathbf{a}, \mathbf{b} \in \R^{3}$. For the ramus corridors, we randomly swap the start and endpoints to simulate the retrograde and antegrade approaches. We then uniformly sample the initial wire tip position within 5\,mm of $\mathbf{a}$ and the direction within $15^{\circ}$ of $\mathbf{b} - \mathbf{a}$.

\para{Sample desired view} The desired view is sampled from views appropriate for the current target corridor. For example, appropriate views for evaluating wire placement in the superior ramus corridor are typically the inlet and obturator oblique views, and other views are sampled with a smaller probability. We refer to the ``oblique left'' and ``oblique right'' view independent of the affected patient side, so that for the right pubic ramus, the obturator oblique is the ``oblique left'' view, and the iliac oblique is ``oblique right.'' We define the ``ideal'' principle ray direction $\raydir^{*}$ for each standard view in the anterior pelvic plane (APP) coordinate system, (see supplement) and the ideal viewing point $\raypoint^{*}$ as the midpoint of the target corridor.
At the beginning of each sequence, we sample the intrinsic camera matrix of the virtual C-arm with sensor width $\sensor \sim \uniform(300, 400)$\,mm, $\sourcedetector \sim \uniform(900, 1200)$, and an image size of $384 \times 384$. Given a viewing point and direction $(\raypoint, \raydir)$, the camera projection matrix $\mathbf{P}$ is computed with the X-ray source (or camera center) at $\raypoint - \sourcepoint \raydir$ and principle ray $\raydir$, where $\sourcepoint ~ \uniform(0.65\,\sourcedetector, 0.75\,\sourcedetector)$ is the source-to-viewpoint distance, and $\sourcedetector$ is the source-to-detector distance (or focal length) of the virtual C-arm.

\para{Evaluate view} Given a current view $(\raypoint, \raydir)$ and desired view $(\raypoint^{*}, \raydir^{*})$, we first evaluate whether the current view is acceptable and, if it is not, make a random adjustment. View evaluation considers the principle ray alignment and whether the viewing point is reasonably centered in the image, computing,
\begin{equation}
  \label{eq:1}
  \raydir \cdot \raydir^{*} < \cos(\theta_{t})
  \text{~~~AND~~~}
  \left|\left|\mathbf{P} \raypoint^{*} - \begin{bmatrix}\frac{H}{2} & \frac{W}{2} & 1\end{bmatrix}^T\right|\right| < \frac{2}{5} \min(H,W)
\end{equation}
where the angular tolerance $\theta_t \in [3^{\circ}, 10^{\circ}]$ depends on the desired view, ranging from teardrop views (low) to lateral (high tolerance).

\para{Sample view} If Eq.~\ref{eq:1} is not satisfied, then we sample a new view $(\raypoint, \raydir)$ uniformly within a uniform window that shrinks every iteration by the adjustment factor, according to
\begin{align}
  \label{eq:2}
    \raypoint & \sim \unifsphere\left(\raypoint^{*}, \clip{\skillfactor \left|\left|\raypoint^{*} - \raypoint\right|\right|}{5\,\rm mm}{100\,\rm mm} \right) \\
    \raydir & \sim \unifangle\left(\raydir^{*}, \clip{\skillfactor \arccos(\raydir^{*} \cdot \raydir)}{1^{\circ}}{45^{\circ}} \right),
\end{align}
where $\unifsphere(\mathbf{c}, r)$ is the uniform distribution in the sphere with center $\mathbf{c}$ and radius $r$, and $\unifangle(\raydir, \theta)$ is the uniform distribution on the solid angle centered on $\raydir$ with colatitude angle $\theta$. This formula emulates observed fluoro-hunting by converging on the desired view until a point, when further adjustments are within the same random window \cite{Killeen2022Dec}. We proceed by alternating view evaluation and sampling until evaluation is satisfied, at which point the simulation resumes with the current activity: wire positioning, wire insertion, or screw insertion.

\para{Evaluate wire placement}
During wire positioning, we evaluate the current wire position and make adjustments from the current view, iterating until evaluation succeeds. Given the current wire tip $\wiretip$, direction $\wiredir$, and projection matrix $\mathbf{P}$, the wire placement is considered ``aligned'' if it \emph{appears} to be aligned with the projected target corridor in the image, modeled as a cylinder. Algorithm 1 (see supplement) details this process for down-the-barrel views (when the principle ray aligns with the target corridor) and orthogonal views. In addition, we include a small likelihood of a false positive evaluation, which diminishes as the wire is inserted.

\para{Sample wire placement} If the wire evaluation determines the current placement is unsuitable, then a new wire placement is sampled. For the down-the-barrel views, this is done similarly to Eq.~\ref{eq:2}, by bringing the wire closer to the corridor in 3D. For orthogonal views, repositioning consists of a small random adjustment to $\wiretip$, a rotation about the principle ray (the in-plane component), and a minor perturbation orthogonal to the ray (out-of-plane). This strategy emulates real repositioning by only adjusting the degree of freedom visible in the image, i.e. the projection onto the image plane:
\begin{align}
  \label{eq:3}
  \wiretip & \sim \unifsphere(\wiretip, \clip{\skillfactor ||\wiretip - \mathbf{a}||}{ 5\,\mathrm{mm}}{10\,\mathrm{mm}})\\
  \wiredir & \gets \rot{\wiredir \times \raydir}{\theta_{\perp}}\rot{\raydir}{\theta^{*} + \theta_{\parallel}} \text{, where }\theta_{\perp} \sim \uniform(-0.1\,\theta^{*}, 0.1\,\theta^{*}),\\
           & ~~~~~~~~~~~~~~~~~~~\theta_{\parallel} \sim \uniform(-\clip{\skillfactor \theta^{*}}{3^{\circ}}{10^{\circ}},~\clip{\skillfactor \theta^{*}}{3^{\circ}}{10^{\circ}}),
\end{align}
and $\theta^{*}$ is the angle between the wire and the target corridor in the image plane. If the algorithm returns ``Good,'' the sequence either selects a new view to acquire (and stays in the \poswire activity) or proceeds to \insertwire or \insertscrew, according to random transitions.


In our experiments, we used 337 CT images: 10 for validation, and 327 for generating the training set. A DRR was acquired at every decision point in the simulation, with a maximum of 1000 images per sequence, and stored along with segmentations and anatomical landmarks. We modeled a K-wire with 2\,mm diameter and orthopedic screws with lengths from 30 to 130\,mm and a 16\,mm thread, with up to eight instances of each in a given sequence. Using a customized version of DeepDRR~\cite{Unberath2018Sep}, we parallelized image generation across 4 RTX 3090 GPUs with an observed GPU memory footprint of $\sim 13$ GB per worker, including segmentation projections. Over approximately five days, this resulted in a training set of 726 sequences totaling 229,488 images and 11 validation sequences with 3,916 images.

\subsection{Transformer Architecture for X-ray-based SPR}
\label{sec:architecture}

Fig.~\ref{fig:architecture} shows the transformer architecture used to predict surgical phases based on embedding tokens for each frame. To encourage local temporal features in each embedding token, we cross-pollinate adjacent frames in the channel dimension, so that each $(3, H, W)$ encoder input contains the previous, current, and next frame. The image encoder is a U-Net~\cite{Ronneberger2015Nov} encoder-decoder variant with 5 Down and Up blocks and 33 spatial output channels, consisting of
\begin{inlistalpha}
  \item 7 segmentation masks of the left hip, right hip, left femur, right femur, sacrum, L5 vertebra, and pelvis;
  \item 8 segmentation masks of bony corridors, including the ramus (2), teardrop (2) and sacrum corridors (4), as in Fig.~\ref{fig:annotations};
  \item 2 segmentation masks for wires and screws; and
  \item 16 heatmaps corresponding to the anatomical landmarks in Fig.~\ref{fig:annotations}.
\end{inlistalpha}
These spatial annotations provide additional supervision, trained with DICE loss $\mathcal{L}_{\rm DICE}$ for segmentation channels and normalized cross correlation $\mathcal{L}_{\rm NCC}$ for heatmap channels as in \cite{Bier2018Sep, GaoSyntheX}. To compute tokens for input to the transformer, we apply a $1 \times 1$ Conv + BatchNorm + ReLU block with kernel size $512$ to the encoder output, followed by global average pooling. The transformer has 6 layers with 8 attention heads and a feedforward dimension of 2048. During training and inference, we apply forward masking so that only previous frames are considered. The output of the transformer are vectors in $\R^{21}$ with phase predictions for each frame, corresponding to
\begin{inlistalpha}
  \item the 8 target corridors;
  \item 3 activities (\poswire, \insertwire, or \insertscrew);
  \item 8 standard views (see Section~\ref{sec:image-seq}); and
  \item 2 frame values (\hunting or \assessment).
\end{inlistalpha}
We compute the cross entropy loss separately for the corridor $\mathcal{L}_{\rm cor}$, activity $\mathcal{L}_{\rm act}$, view $\mathcal{L}_{\rm view}$, and frame $\mathcal{L}_{\rm fr}$ phases, and take the mean. See the supplement for more training details.

\section{Evaluation}
\label{sec:results}
\para{Simulation}
We report the results of our approach first on simulated image sequences, generated from the withheld set of CT images, which serves as an upper bound on real X-ray performance. In this context our approach achieves an accuracy of 96.9\%, 86.3\%, 93.9\%, and 98.2\% with respect to the corridor, activity, view, and frame level, respectively, for an average of 93.8\% across all levels. We observe comparatively lower performance for the activity level and speculate that this occurs because the \insertwire activity visually resembles \poswire for low insertions, in our simulation. Moreover, we achieve an average DICE score of 0.73 and landmark detection error of $1.01 \pm 0.153$ pixels in simulation, indicating that these features provide a meaningful signal.

\begin{figure}[t]
    \centering
    \includegraphics[width=\columnwidth]{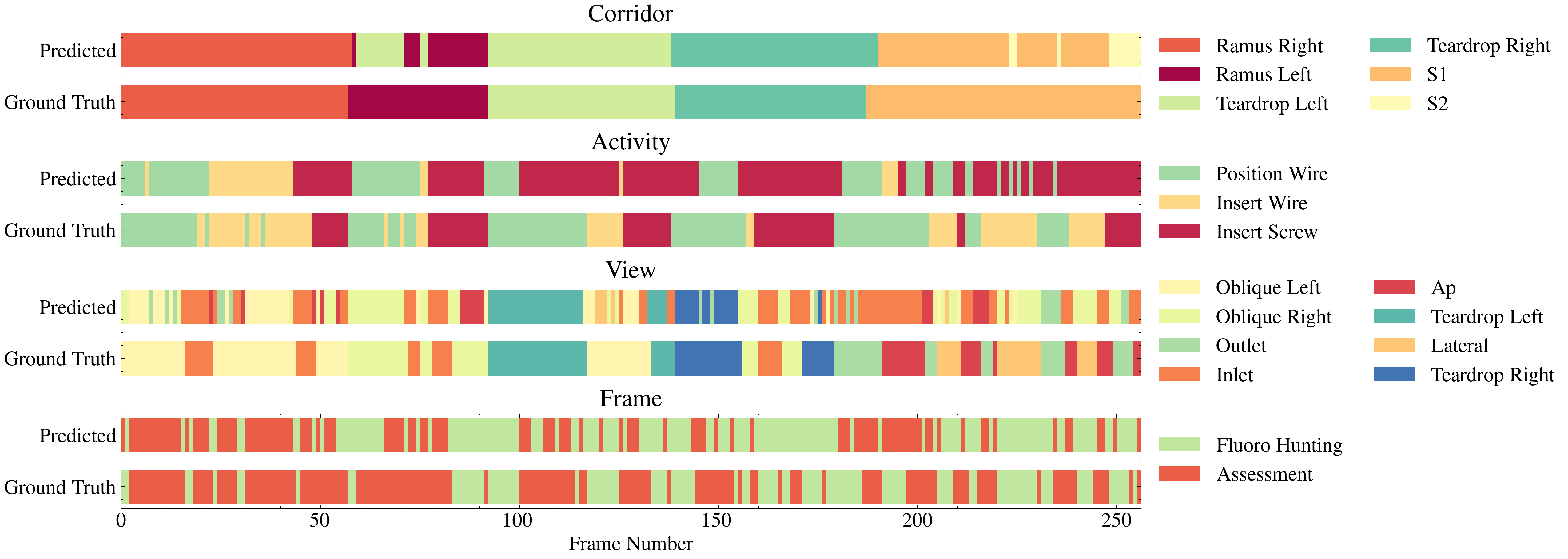}
    \caption{Results of surgical phase recognition for a cadaveric procedure. We observe varying performance based on the target corridor, either because of the associated views or due to the accumulated orthopedic hardware.}
    \label{fig:cadaver-results}
\end{figure}

\para{Cadaver study}
We evaluate our approach on cadaveric image sequences with five wire and screw insertions. An attending orthopedic surgeon performed percutaneous fixation on a lower torso specimen, taking the antegrade approach for the left and right pubic rami corridors, followed by the left and right teardrop and S1 screws. An investigator acted as the radiological technician, manipulating a Siemens CIOS Fusion C-arm according to the surgeon's direction. A total of 257 images were acquired during these fixations. Two investigators recorded phase labels during the procedure based on the surgeon's input. Although segmentation masks and anatomical landmarks are not available for these images, we observe qualitatively satisfactory segmentation masks (see supplement), indicating successful sim-to-real generalization.

Our results for phase recognition, shown in Fig.~\ref{fig:cadaver-results} demonstrate the potential for Pelphix as a viable approach to SPR in X-ray. We achieve an overall accuracy of 88\%, 61\%, 51\%, and 70\% with respect to the corridor, activity, view, and frame levels, respectively. 
We find that across all levels, accuracy varied significantly depending on the target corridor, likely because of the associated views. For instance, prediction of the right ramus, left teardrop, and right teardrop corridors was achieved with 100\%, 98\%, and 100\% accuracy, while the left ramus and S1 corridors yielded 57\% and 80\% accuracy, respectively. Similar variation can be seen in the activity, acquisition, and frame level accuracy: screw insertion was recognized with nearly 100\% accuracy, while wire insertion was often confused for screw insertion. This is because our simulation varied the screw insertion depth randomly rather than based on the anatomy. Teardrop and inlet views are recognized with reasonable accuracy (90\%, 60\%, and 81\%), while the network struggles with lateral views. These shortcomings may reflect sampling biases in the stochastic simulation that resulted in certain views being underrepresented, but the fact that the least represented views are the left and right teardrops (3.5\% and 2.3\% of images) would seem to discount this.

\section{Discussion and Conclusion}

As our results show, Pelphix is a potentially viable approach to robust SPR based on X-ray images. We showed that stochastic simulation of percutaneous fracture fixation, despite having no access to real image sequences, is a sufficiently realistic data source enabling sim-to-real transfer. While we expect adjustments to the simulation approach will close the gap even further, truly performative SPR algorithms for X-ray may rely on Pelphix-style simulation for pretraining, before fine-tuning on real image sequences to account for human-like behavior. Extending this approach to other procedures in orthopedic surgery, angiography, and interventional radiology will require task-specific simulation capable of modeling possibly more complex tool-tissue interactions and human-in-the-loop workflows. Nevertheless, Pelphix provides a first viable route toward X-ray-based surgical phase recognition, which we hope will motivate routine collection and interpretation of these data, in order to enable advances in surgical data science that ultimately improve the standard of care for patients.

\bibliographystyle{splncs04}
\bibliography{references}

\end{document}